\documentclass{article} 
\usepackage{iclr2017_conference,times}
\usepackage{url}

\usepackage{mathtools}
\usepackage{amsfonts}
\usepackage{graphicx}

\graphicspath{{./figures/}}

\usepackage{multirow}
\usepackage{siunitx}
\usepackage{subcaption}

\usepackage{hyperref}

\newcommand\entvec[1]{\boldsymbol{v}_{\texttt{#1}}}

\newcommand\relcov[1]{\boldsymbol{\Sigma}_{\texttt{#1}}}

\newcommand{\RR}{\mathbb{R}}
\DeclareMathOperator*{\EE}{\mathbb{E}}

\newcommand{\vb}{\boldsymbol{b}}
\newcommand{\vv}{\boldsymbol{v}}
\newcommand{\vw}{\boldsymbol{w}}
\newcommand{\vh}{\boldsymbol{h}}
\newcommand{\vo}{\boldsymbol{o}}
\newcommand{\vmu}{\boldsymbol{\mu}}
\newcommand{\vdelta}{\boldsymbol{\delta}}
\newcommand{\mSigma}{\boldsymbol{\Sigma}}
\newcommand{\mM}{\boldsymbol{M}}
\newcommand{\mW}{\boldsymbol{W}}
\newcommand{\vu}{\boldsymbol{u}}

\newcommand{\playerblank}{\underline{(player)}}
\newcommand{\countryblanka}{\underline{(country\_1)}}
\newcommand{\countryblankb}{\underline{(country\_2)}}
\newcommand{\clubblank}{\underline{(club)}}

\newcommand{\positionblank}{\underline{(position)}}

\title{Gaussian Attention Model and Its Application to Knowledge Base Embedding and
Question Answering}

\author{Liwen Zhang \\
Department of Computer Science\\
University of Chicago\\
Chicago, IL 60637, USA\\
\texttt{liwenz@cs.uchicago.edu}
\And
John Winn \& 
Ryota Tomioka \\
Microsoft Research Cambridge \\
Cambridge, CB1 2FB, UK\\
\texttt{\{jwinn, ryoto\}@microsoft.com}
}

\usepackage{amsmath,amssymb}	
\begin{document}

\maketitle

\begin{abstract}
We propose the Gaussian attention model for content-based neural memory
access. With the proposed attention model, a neural network has the
additional degree of freedom to control the focus of its attention from
a laser sharp attention to a broad attention. It is applicable whenever
we can assume that the distance in the latent space reflects some notion
of semantics. We use the proposed attention model as a scoring function
for the embedding of a knowledge base into a continuous vector space and
then train a model that performs question answering about the entities
in the knowledge base. The proposed attention model can handle both the
propagation of uncertainty when following a series of relations and also
the conjunction of conditions in a natural way. On a dataset of soccer
players who participated in the FIFA World Cup 2014, we demonstrate that
our model can handle both path queries and conjunctive queries well.
\end{abstract}

\section{Introduction}

There is a growing interest in incorporating external memory into
neural networks. For example, memory networks
\citep{weston2014memory,sukhbaatar2015end} are equipped with static memory
slots that are content or location addressable. Neural Turing machines
\citep{graves2014neural} implement memory slots that can be read and
written as in Turing machines \citep{turing1938computable} but through
differentiable attention mechanism.

Each memory slot in these models stores a vector corresponding to a
continuous representation of the memory content.
In order to recall a piece of information stored in memory,
attention is typically employed. Attention mechanism introduced by
\cite{bahdanau2014neural} uses a network that outputs
a discrete probability mass over memory items.
A memory read can be
implemented as a weighted sum of the memory vectors in which the weights
are given by the attention network. Reading out a single item can be
realized as a special case in which the output of the attention network
is peaked at the desired item.
The attention network may depend on the
current context as well as the memory item itself.
The attention model is called location-based and
content-based, if it depends on the location in the memory and the stored
memory vector, respectively.

Knowledge bases, such as WordNet and Freebase, can also be stored in
memory either through an explicit knowledge base embedding
\citep{bordes2011learning,nickel2011three,socher2013reasoning} or
through a feedforward network \citep{bordes2015large}.

When we embed entities from a knowledge base in a continuous vector
space, if the capacity of the embedding model is appropriately controlled,
we expect semantically similar entities to be close to
each other, which will allow the model to generalize to unseen
facts. However the notion of proximity may strongly depend on the type
of a relation. For example, Benjamin Franklin was an engineer but also a
politician. We would need different metrics to capture his proximity to
other engineers and politicians of his time.

In this paper, we propose a new attention model for content-based
addressing. Our model scores each item $\entvec{item}$ in the memory by
the (logarithm of) multivariate Gaussian likelihood as follows:
\begin{align}
 \text{score}(\entvec{item}) &= \log
 \phi(\entvec{item}|\vmu_{\texttt{context}}, \relcov{context})\notag\\
\label{eq:gaussian-score}
  &=-\frac{1}{2}(\entvec{item}-\vmu_{\texttt{context}})\relcov{context}^{-1}(\entvec{item}-\vmu_{\texttt{context}})+\text{const.}
\end{align}
where \texttt{context} denotes all the variables that the attention
depends on. For example, ``American engineers in the 18th century'' or
``American politicians in the 18th century'' would be two contexts that
include Benjamin Franklin but the two attentions would have very different shapes.

\begin{figure}[tb]
 \begin{center}
  \includegraphics[width=.5\textwidth]{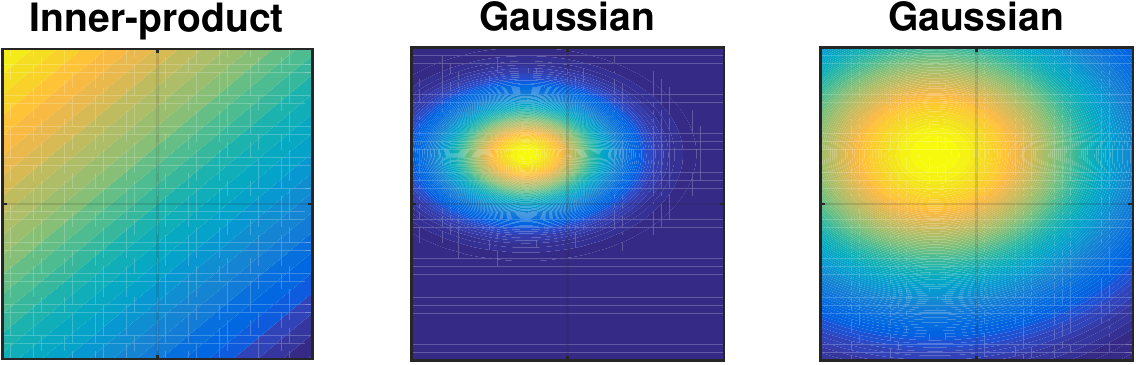}
  \caption{Comparison of the conventional content-based attention model using
  inner product and the proposed Gaussian attention model with the same
  mean but two different covariances.}
  \label{fig:linquad}
 \end{center}
\end{figure}

Compared to the (normalized) inner product used in previous work
\citep{sukhbaatar2015end,graves2014neural} for content-based addressing,
the Gaussian model has the additional control of the spread of the
attention over items in the memory. As we show in Figure
\ref{fig:linquad}, we can view the conventional inner-product-based
attention and the proposed Gaussian attention
as addressing by an affine energy function and a quadratic energy
function, respectively. By making the addressing mechanism more complex,
we may represent many entities in a relatively low dimensional embedding
space. Since knowledge bases are typically extremely sparse, it is more
likely that we can afford to have a more complex attention model than a
large embedding dimension.

We apply the proposed Gaussian attention model to question
answering based on knowledge bases. At the high-level, the goal of the
task is to learn the mapping from a question about objects in the
knowledge base in natural language to a probability distribution over the
entities. We use the scoring function \eqref{eq:gaussian-score} for both
embedding the entities as vectors, and extracting the conditions
mentioned in the question and taking a conjunction of them to score each
candidate answer to the question.

The ability to compactly represent a set of objects makes the
Gaussian attention model well suited for
representing the uncertainty in a multiple-answer question (e.g.,
``who are the children of Abraham Lincoln?''). Moreover, traversal over the
knowledge graph~\citep[see][]{guu2015traversing} can be naturally
handled by a series of Gaussian convolutions, which
generalizes the addition of vectors. In fact, we model each relation as
a Gaussian with mean and variance parameters. Thus a traversal on a
relation corresponds to a translation in the mean and addition of
the variances.

\begin{figure}[tb]
 \begin{center}
  \fbox{\includegraphics[clip,width=\textwidth]{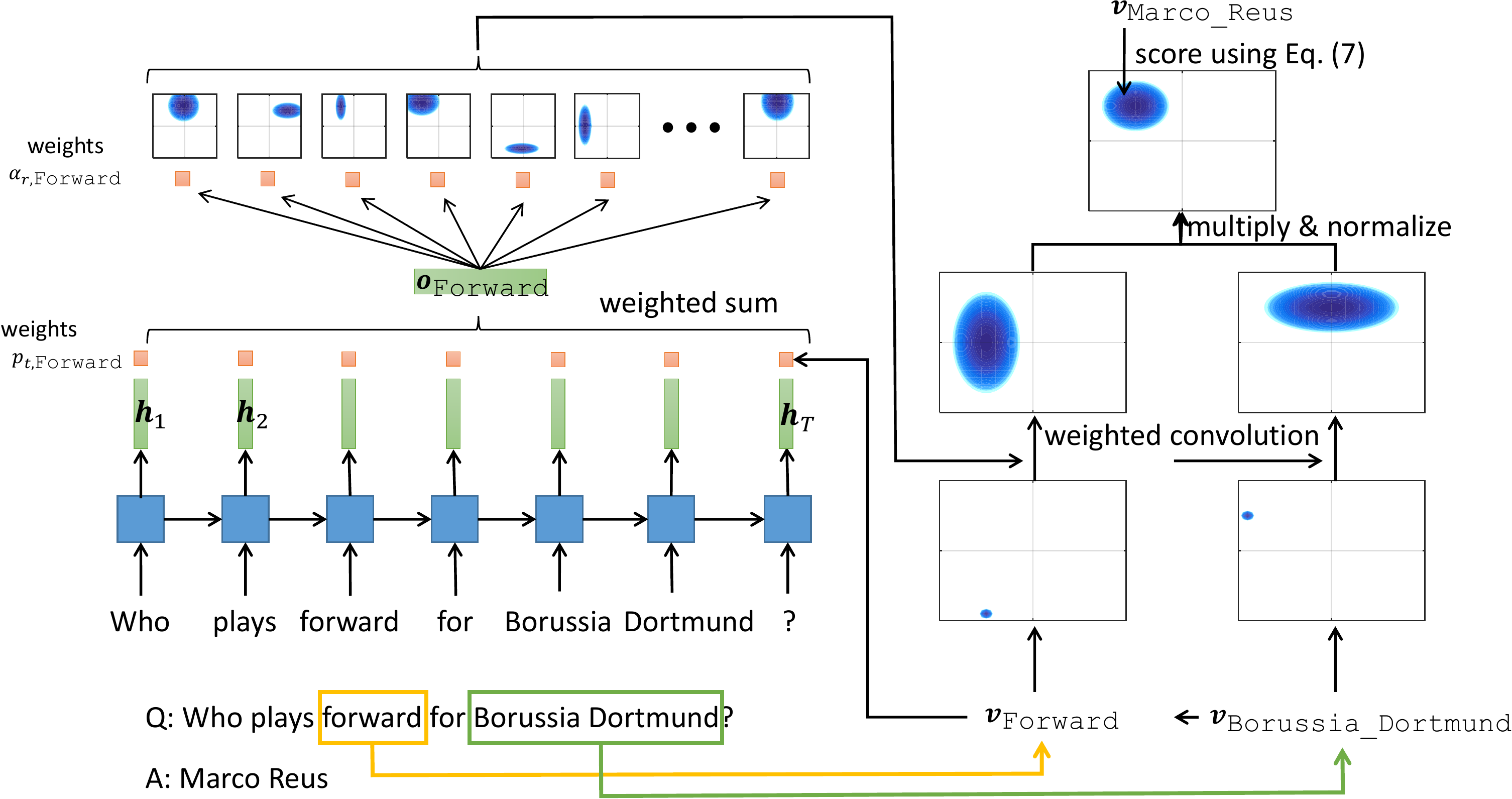}}
  \caption{
  The input to the system is a question in natural language.
  Two entities 
  \texttt{Forward} and \texttt{Borussia\_Dortmund} are identified
  in the question and associated with point mass distributions centered
  at the corresponding entity
  vectors. An LSTM encodes the input into a sequence of output
  vectors of the same length. Then we take average of the
  output vectors weighted by attention $p_{t,e}$ for each recognized
  entity $e$ to predict the weight $\alpha_{r,e}$ for relation $r$
  associated with entity $e$. We form a Gaussian attention over the
  entities for each entity $e$ by convolving the corresponding point
  mass with the (pre-trained) Gaussian embeddings of the relations
  weighted by $\alpha_{r,e}$ according to Eq.~\eqref{eq:mean-cov-qa}. The final prediction is produced by taking
  the product and normalizing the Gaussian attentions.
  }
  \label{fig:flow}
 \end{center}
\end{figure}

The proposed question answering model is able to handle not only
the case where the answer to a question is associated with an atomic
fact, which is called simple Q\&A \citep{bordes2015large}, but also
questions that require composition of relations (path queries in
\cite{guu2015traversing}) and conjunction of queries. An example flow
of how our model deals with a question ``Who plays forward for Borussia
Dortmund?'' is shown in Figure \ref{fig:flow} in Section \ref{sec:qanda}.

This paper is structured as follows. In Section \ref{sec:embedding}, we
describe how the Gaussian scoring function \eqref{eq:gaussian-score} can
be used to embed the entities in a knowledge base into a continuous
vector space. We call our model TransGaussian because of its similarity
to the TransE model proposed by \cite{bordes2013translating}. Then in Section
\ref{sec:qanda}, we describe our question answering model.
In Section \ref{sec:experiments}, we carry out experiments on
WorldCup2014 dataset we collected. The dataset is relatively small but
it allows us to evaluate not only simple questions but also path queries
and conjunction of queries. The proposed
TransGaussian embedding with the question answering model achieves
significantly higher accuracy than the vanilla TransE embedding or
TransE trained with compositional relations \cite{guu2015traversing}
combined with the same question answering model.

\section{Knowledge Base embedding}
\label{sec:embedding}
In this section, we describe the proposed TransGaussian model based on 
the Gaussian attention model \eqref{eq:gaussian-score}.
While it is possible to train a network that computes
the embedding in a single pass \citep{bordes2015large} or over
multiple passes \citep{li2015gated}, it is more efficient
to offload the embedding as a separate step for question
answering based on a large static knowledge base.

\subsection{The TransGaussian model}
Let $\mathcal{E}$ be the set of entities and $\mathcal{R}$ be the set of
relations. A knowledge base is a collection of triplets $(s,r,o)$, where
we call $s\in\mathcal{E}$,  $r\in\mathcal{R}$, and $o\in\mathcal{E}$,
the subject, the relation, and the object of the triplet,
respectively. Each triplet encodes a {\em fact}. For example,
$(\texttt{Albert\_Einstein},\texttt{has\_profession},\texttt{theoretical\_physicist})$.
%
%
All the triplets given in a knowledge base are assumed to be true. However
generally speaking a triplet may be true or false. Thus
knowledge base embedding aims at training a model that predict if a
triplet is true or not given some parameterization of the entities and
relations
\citep{bordes2011learning,bordes2013translating,nickel2011three,socher2013reasoning,wang2014knowledge}. 

In this paper, we associate a vector $\vv_s \in \mathbb{R}^d$ with each
entity $s \in \mathcal{E}$, and we associate each relation $r \in
\mathcal{R}$ with two parameters, $\vdelta_r \in \mathbb{R}^d$
and a positive definite symmetric matrix $\mSigma_r \in \mathbb{R}^{d\times d}_{++}$.

Given subject $s$ and relation $r$, we can compute the score of an object
$o$ to be in triplet $(s,r,o)$ using the Gaussian attention model as
\eqref{eq:gaussian-score} with 
\begin{align}
  \label{eq:triplet-score}
  \text{score}(s,r,o)&=\log\phi(\vv_o|\vmu_{\texttt{context}},\relcov{context}),
\end{align}
where $\vmu_{\texttt{context}} = \vv_s + \vdelta_r$, $\relcov{context} = \mSigma_r$.
Note that if $\mSigma_r$ is fixed to the identity matrix, we are
modeling the relation of subject $\vv_s$ and object $\vv_o$ as a
translation $\vdelta_r$, which is
equivalent to the TransE model \citep{bordes2013translating}. We allow
the covariance $\mSigma_r$ to depend on the relation to handle
one-to-many relations (e.g., \texttt{profession\_has\_person} relation)
and capture the shape of the distribution of the set of objects that can
be in the triplet. We call our model \textbf{TransGaussian} because of its similarity
to TransE  \citep{bordes2013translating}.

\paragraph{Parameterization} For computational efficiency, we will
restrict the covariance matrix $\mSigma_r$ to be diagonal in this
paper. Furthermore, in order to ensure that $\mSigma_r$ is strictly
positive definite, we employ the  exponential linear unit \citep[ELU,][]{clevert2015fast} and parameterize $\mSigma_r$ as follows:
\begin{align*}
 \mSigma_r = {\rm diag}
 \left(
\begin{smallmatrix}
 {\rm ELU}(m_{r,1})+1+\epsilon & & \\
  & \ddots & \\
  & & {\rm ELU}(m_{r,d})+1+\epsilon
\end{smallmatrix}
 \right)
\end{align*}
where $m_{r,j}$ ($j=1,\ldots,d$) are the unconstrained parameters that are
optimized during training and $\epsilon$ is a small positive value 
that ensure the positivity of the variance during numerical computation.  The ELU is defined as
\begin{align*}
{\rm ELU}(x) = \begin{cases}
x , & x \geq 0, \\
\exp{(x)}-1, & x < 0.
\end{cases}
\end{align*}

\paragraph{Ranking loss}
Suppose we have a set of triplets $ \mathcal{T} = \left\{ \left( s_i,
r_i, o_i \right )\right\}_{i=1}^{N}$ from the knowledge base. Let
$\mathcal{N}( s, r) $ be the set of incorrect objects to be in the triplet $(s, r, \cdot)$.

Our objective function uses 
the ranking loss to measure the margin between the scores of true answers and
those of false answers and it can be written as follows:
 \begin{align}
\min_{ \substack{ \{\vv_e : e\in \mathcal{E}\}, \\ \{\vdelta_r, \mM_r,: r\in \bar{\mathcal{R}}\}}}
 \frac{1}{N} \sum_{(s,r,o)\in\mathcal{T}} &\mathbb{E}_{t' \sim \mathcal{N}(s,r)}\left[
\left[ \mu -
 \text{score}(s, r, o) + \text{score}(s, r, t') \right]_+\right]\notag\\
  \label{eq:objective-simple}
  &+\lambda\left( \sum_{e\in\mathcal{E}}\|\vv_e\|_2^2+ \sum_{r\in \bar{\mathcal{R}}} \left( \|\vdelta_r\|_2^2+\|\mM_r\|_{F}^2 \right)\right),
 \end{align}
where, $N=|\mathcal{T}|$, $\mu$ is the margin parameter and $\mM_r$ denotes the diagonal
matrix with $m_{r,j}$, $j=1,\ldots,d$ on the diagonal;
the function $[\cdot]_+$ is defined as $[x]_+=\max(0,x)$.
Here, we treat an inverse relation as a separate relation and denote by
$\bar{\mathcal{R}}=\mathcal{R}\cup\mathcal{R}^{-1}$ the set of all the
relations including both relations in $\mathcal{R}$ and their inverse
relations; a relation $\tilde{r}$ is the inverse relation of $r$ if
$(s,\tilde{r},o)$ implies $(o,r,s)$ and vice versa.
Moreover, $\mathbb{E}_{t' \sim \mathcal{N}(s,r)}$
denotes the expectation with respect to the uniform distribution over
the set of incorrect objects, which we approximate with 10 random
samples in the experiments. Finally, the last terms are $\ell_2$
regularization terms for the embedding parameters.

\subsection{Compositional relations}
\label{sec:comp-rel}
\citet{guu2015traversing} has recently shown that
training TransE with {\em compositional relations} can make it competitive
to more complex models, although TransE is much simpler compared
to for example, neural tensor networks (NTN, \citet{socher2013reasoning}) and TransH
\cite{wang2014knowledge}.
Here, a compositional relation is a
relation that is composed as a series of relations in $\mathcal{R}$, for
example, $\texttt{grand\_father\_of}$ can be composed as first applying
the $\texttt{parent\_of}$ relation and then the $\texttt{father\_of}$
relation, which can be seen as a traversal over a path on the knowledge graph.

TransGaussian model can naturally handle and propagate the uncertainty
over such a chain of relations by convolving the Gaussian distributions
along the path. That is, the score of an entity $o$ to be in the $\tau$-step relation
$r_1/r_2/\cdots/r_\tau$ with subject $s$, which we denote by 
the triplet $(s,r_1/r_2/\cdots/r_\tau,o)$, is given as
 \begin{align}
  \label{eq:score-comp}
 \text{score}(s,r_1/r_2/\cdots/r_\tau,o)&=\log\phi(\vv_o|\vmu_{\texttt{context}},\relcov{context}),
 \end{align}
 with $\vmu_{\texttt{context}} = \vv_s + \sum_{t=1}^{\tau}\vdelta_{r_t}$,
 $\relcov{context} = \sum_{t=1}^{\tau}\mSigma_{r_t}$,
where the covariance associated with each relation is parameterized in the same way as in the previous subsection.

\paragraph{Training with compositional relations}
Let $\mathcal{P}=\left\{ \left( s_i, r_{i_1} / r_{i_2} / \cdots /
r_{i_{l_i}} , o_i \right) \right\}_{i=1}^{N'}$ be a set of randomly 
sampled paths from the knowledge graph. Here relation $r_{i_k}$ in a
path can be a relation in $\mathcal{R}$ or an inverse relation in $\mathcal{R}^{-1}$.
With the scoring function \eqref{eq:score-comp}, the generalized
training objective for compositional relations can be written 
identically to 
\eqref{eq:objective-simple} except for replacing $\mathcal{T}$ with
$\mathcal{T}\cup\mathcal{P}$ and replacing $N$ with $N'=|\mathcal{T}\cup\mathcal{P}|$.

\section{Question answering}
\label{sec:qanda}
Given a set of question-answer pairs, in which the question is phrased
in natural language and the answer is an entity in the knowledge base,
our goal is to train a model that learns the mapping from the question
to the correct entity.
Our question answering model consists of three steps, entity
recognition, relation composition, and conjunction.
We first identify a list of entities mentioned in the
question (which is assumed to be provided by an oracle in this paper).  If the question is ``Who plays
Forward for Borussia Dortmund?'' then the list would be
[\texttt{Forward}, \texttt{Borussia\_Dortmund}].
The next step is to predict the path of relations on the
knowledgegraph starting from each entity in the list extracted in
the first step. In the above example, this will be
(smooth versions of)
\texttt{/Forward/position\_played\_by/} and \texttt{/Borussia\_Dortmund/has\_player/}
predicted as series of Gaussian convolutions.
In general, we can have multiple relations appearing
in each path.
Finally, we take a product of all the Gaussian
attentions and renormalize it, which is
equivalent to Bayes' rule with independent observations (paths) and a noninformative prior.

\subsection{Entity recognition}
We assume that there is an oracle that provides a list containing all the entities
mentioned in the question, because (1) a domain specific entity recognizer can be 
developed efficiently \citep{williams2015fast} and (2) generally entity recognition
is a challenging task and it is beyond the scope of this paper to show whether there
is any benefit in training our question answering model jointly with a entity recognizer.
We assume that the number of extracted entities can be 
different for each question.


\subsection{Relation composition}
We train a long short-term memory \citep[LSTM,][]{hochreiter1997long} network that emits an output $\vh_t$ for each token in the input sequence. Then we compute the attention over the hidden states for each recognized entity $e$ as
\begin{align*}
 p_{t,e} = \textrm{softmax}\left(f(\vv_e, \vh_t)\right)\quad(t=1,\ldots,T),
\end{align*}
where $\vv_e$ is the vector associated with the entity $e$.
We use a two-layer perceptron for $f$ in our experiments, which can be
written as follows:
\begin{align*}
f(\vv_e,\vh_t) &= \vu_f^\top \text{ReLU} \left( \mW_{f,v}\vv_e+
\mW_{f,h}\vh_t + \vb_1
 \right)+b_2,
\end{align*}
where $\mW_{f,v}\!\in\!\RR^{L\times d}$, $\mW_{f,h}\!\in\!\RR^{L\times H}$,
$\vb_1\in\RR^{L}$, $\vu_f\in\RR^{L}$, $b_2\in\RR$ are parameters. Here ${\rm ReLU}(x)\!=\!\max(0,x)$ is the rectified linear unit.
Finally, softmax denotes softmax over the $T$ tokens.

Next, we use the weights $p_{t,e}$ to compute the weighted sum over the hidden states $\vh_t$ as
\begin{align}
\label{eq:attention-weights}
 \vo_e = \sum\nolimits_{t=1}^{T}p_{t,e} \vh_t.
\end{align}

Then we compute the weights $\alpha_{r,e}$ over all the relations as
$
\alpha_{r,e} = {\rm ReLU}\left(\vw_{r}^\top \vo_e\right)\quad (\forall r\in\mathcal{R}\cup\mathcal{R}^{-1}).
$
Here the rectified linear unit is used to ensure the positivity of the
weights. Note however that the 
weights should not be normalized, because we may want to use the same
relation more than once in the same path. Making the weights positive
also has the effect of making the attention sparse and interpretable
because there is no cancellation.

For each extracted entity $e$, we view the extracted entity and the answer of the question
to be the subject and the object in some triplet
$(e,p,o)$, respectively, where the path $p$ is inferred from the question
as the weights $\alpha_{r,e}$ as we described above. Accordingly, the
score for each candidate answer $o$ can be expressed using
\eqref{eq:gaussian-score} as:
\begin{align}
 \label{eq:mean-cov-qa}
 \text{score}_e(\vv_{o})=&\log
 \phi(\vv_o|\vmu_{e,\alpha,\text{KB}},\mSigma_{e,\alpha,\text{KB}})
\end{align}
with
$\vmu_{e,\alpha,\text{KB}}=\vv_e
 +\sum_{r\in\bar{\mathcal{R}}}\alpha_{r,e}\vdelta_r$,
$\mSigma_{e,\alpha,\text{KB}}=\sum_{r\in\bar{\mathcal{R}}}\alpha_{r,e}^2\mSigma_r$,
where $\vv_e$ is the vector associated with entity $e$ and
$\bar{\mathcal{R}}=\mathcal{R}\cup\mathcal{R}^{-1}$ denotes the set of
relations including the inverse relations.

\subsection{Conjunction}
Let $\mathcal{E}(q)$ be the set of entities recognized in the question $q$. The final
step of our model is to take the conjunction of the Gaussian attentions
derived in the previous step. This step is simply carried out by
multiplying the Gaussian attentions as follows:
\begin{align}
 \text{score}(\vv_o\vert \mathcal{E}(q), \Theta) &=\log \prod_{e\in \mathcal{E}(q)}
 \phi(\vv_o|\vmu_{e,\alpha,\text{KB}},
 \mSigma_{e,\alpha,\text{KB}})\notag\\
 \label{eq:gaussian-conjunction}
 &=-\frac{1}{2}\sum_{e\in
 \mathcal{E}(q)}\left(\vv_o-\vmu_{e,\alpha,\text{KB}}\right)^\top\mSigma_{e,\alpha,\text{KB}}^{-1}(\vv_o-\vmu_{e,\alpha,\text{KB}})+\text{const.},
\end{align}
which is again a (logarithm of) Gaussian scoring function, where $\vmu_{e,\alpha,\text{KB}}$ and $\mSigma_{e,\alpha,\text{KB}}$
are the mean and the covariance of the Gaussian attention given in
\eqref{eq:mean-cov-qa}. Here $\Theta$ denotes all the parameters of the
question-answering model.

\subsection{Training the question answering model}

Suppose we have a knowledge base $\left(\mathcal{E}, \mathcal{R}, \mathcal{T}\right)$
and a trained TransGaussian model 
$\left( \left\{v_e \right\}_{e\in \mathcal{E}}, \left\{ (\vdelta_r ,
\mSigma_r) \right\}_{r \in \bar{\mathcal{R}}}\right)$, where
$\bar{\mathcal{R}}$ is the set of all relations including the inverse relations.
During training time, we assume the training set is a supervised question-answer pairs
$\left\{ \left( q_i, \mathcal{E}(q_i), a_i \right) : i=1,2,\dots,m \right\}$.  
Here, $q_i$ is a question formulated in natural language,
$\mathcal{E}(q_i) \subset \mathcal{E} $ is a set of knowledge base entities that
appears in the question, and $a_i \in \mathcal{E}$ is the answer to the question.
For example, on a knowledge base of soccer players, a valid training sample could be

\noindent
(``Who plays forward for Borussia Dortmund?'',$\left[ \texttt{Forward}, \texttt{Borussia\_Dortmund} \right]$,~\texttt{Marco\_Reus}).

Note that the answer to a question is not necessarily unique and we
allow $a_i$ to be any of the true answers in the knowledge base.
During test time, our model is shown $(q_i, \mathcal{E}(q_i))$ and the task is to
find $a_i$. We denote the set of answers to $q_i$ by $A(q_i)$.

To train our question-answering model, we minimize the objective function
\begin{align*}
\frac{1}{m} \sum_{i=1}^m \Bigl(\EE_{t' \sim \mathcal{N}(q_i) }
&\left[ \left[\mu 
- \text{score}(\vv_{a_i} \vert \mathcal{E}(q_i) , \Theta )
+ \text{score}(\vv_{t'} \vert \mathcal{E}(q_i) , \Theta )
\right]_+ \right] + 
\nu \!\!\!\sum_{e \in \mathcal{E}(q_i)} \sum_{r\in \bar{\mathcal{R}}}\left| \alpha_{r,e}\right|
\Bigr) + \lambda \|\Theta\|_2^2 
\end{align*}
where $\mathbb{E}_{t' \sim \mathcal{N}(q_i) }$ is expectation with respect to a uniform
distribution over of all incorrect answers to $q_i$, which we approximate with 10 random samples.
We assume that the number of relations implied in a question is small compared to
the total number of relations in the knowledge base.  
Hence the coefficients $\alpha_{r,e}$ computed for each question $q_i$ are
regularized by their $\ell_1$ norms.

\section{Experiments}
\label{sec:experiments}

As a demonstration of the proposed framework, 
we perform question and answering on a dataset of soccer players.
In this work, we consider two types of questions. A {\em path query} is a
question that contains only one named entity from the knowledge base 
and its answer can be found from the knowledge graph by walking down a
path consisting of a few relations. A {\em conjunctive query}
is a question that contains more than one entities and the answer is
given as the conjunction of all path queries starting from each entity.
Furthermore, we experimented on a knowledge base completion task with TransGaussian
embeddings to test its capability of generalization to unseen fact.
Since knowledge base completion is not the main focus of this work,
 we include the results in the Appendix. 

\subsection{WorldCup2014 Dataset}
We build a knowledge base of football players that participated in FIFA
World Cup 2014
\footnote{The original dataset can be found at https://datahub.io/dataset/fifa-world-cup-2014-all-players.}.
The original dataset consists of players' information such as nationality, 
positions on the field and ages etc.
We picked a few attributes and constructed 1127 entities and 6 atomic relations.
The entities include 736 players, 297 professional soccer clubs, 51 countries, 39
numbers and 4 positions. And the six atomic relations are
\begin{tabular}[h]{p{.52\textwidth}p{.48\textwidth}}
\texttt{plays\_in\_club}: PLAYER $\rightarrow$ CLUB, &
\texttt{plays\_position}: PLAYER $\rightarrow$ POSITION, \\
\texttt{is\_aged}: PLAYER $\rightarrow$ NUMBER, &
\texttt{wears\_number} \footnote{This is players' jersey numbers in the national teams.}:
PLAYER $\rightarrow$ NUMBER, \\
\texttt{plays\_for\_country}: PLAYER $\rightarrow$ COUNTRY, &
\texttt{is\_in\_country}: CLUB $\rightarrow$ COUNTRY,
\end{tabular}
where PLAYER, CLUB, NUMBER, etc, denote the type of entities that can
appear as the left or right argument for each relation. Some relations
share the same type as the right argument, e.g.,
\texttt{plays\_for\_country} and \texttt{is\_in\_country}. 

Given the entities and relations, we transformed the dataset into a set of
3977 triplets. A list of sample triplets can be found in the Appendix.
Based on these triplets, we created two sets of question answering tasks 
which we call \emph{path query} and \emph{conjunctive query} respectively.
The answer of every question is always an entity in the knowledge base
and a question can involve one or two triplets.  
The questions are generated as follows.

\paragraph*{Path queries.}
Among the paths on the knowledge graph, there are some natural composition of relations, e.g., 
\texttt{plays\_in\_country} (PLAYER $\rightarrow$ COUNTRY) can be decomposed as the
composition of \texttt{plays\_in\_club} (PLAYER$\rightarrow$ CLUB) and 
\texttt{is\_in\_country} (CLUB $\rightarrow$ COUNTRY).
In addition to the atomic relations, we manually picked a few meaningful compositions of relations
and formed {\em query templates}, which takes the form ``find $e\in\mathcal{E}$, such that
$(s, p, e)$ is true'', where $s$ is the subject and $p$ can be an atomic relation or a path of relations.
To formulate a set of path-based question-answer pairs, we manually created one or more
{\em question templates} for every query template (see Table~\ref{tab:query-templates})
Then, for a particular instantiation of a query template with subject and object entities,
we randomly select a question template to generate a question given the subject; the object entity
becomes the answer of the question.
See Table~\ref{tab:path-query-samples} for the list of composed relations, sample questions, and answers.
Note that all atomic relations in this dataset are many-to-one 
while these composed relations can be one-to-many or many-to-many as well.

\paragraph*{Conjunctive queries.}
To generate question-and-answer pairs of conjunctive queries, we first picked three pairs of relations
and used them to create query templates of the form
``Find $e \in \mathcal{E}$, such that both $(s_1, r_1, e)$ and $(s_2, r_2, e)$ are true.''
(see Table \ref{tab:query-templates}).
For a pair of relations $r_1$ and $r_2$, we enumerated all pairs of entities $s_1$, $s_2$ 
that can be their subjects
and formulated the corresponding query in natural language using question templates
as in the same way as path queries. 
See Table~\ref{tab:conjunction-query-samples} for a list of sample questions and answers.

%
As a result, we created 8003 question-and-answer pairs of path queries and 2208 pairs of conjunctive queries which
are partitioned into train / validation / test subsets. 
We refer to Table~\ref{tab:data-stats} for more statistics about the dataset.
Templates for generating the questions are list in Table~\ref{tab:query-templates}.

\subsection{Experimental setup}
To perform question and answering under our proposed framework, we first train the TransGaussian
model on WorldCup2014 dataset.
In addition to the atomic triplets, we randomly sampled 50000 paths with length 1 or 2 from the knowledge graph
and trained a TransGaussian model compositionally as described in Set~\ref{sec:comp-rel}.
An inverse relation is treated as a separate relation.
Following the naming convention from \cite{guu2015traversing}, 
we denote this trained embedding by \emph{TransGaussian (COMP)}.
%
We found that the learned embedding possess some interesting properties.  
Some dimensions of the embedding space dedicate to represent a particular relation. 
Players are clustered by their attributes when entities' embeddings are projected to 
the corresponding lower dimensional subspaces.  
We elaborate and illustrate such properties in the Appendix.

\paragraph*{Baseline methods}
We also trained a TransGaussian model only on the atomic triplets 
and denote such a model by \emph{TransGaussian (SINGLE)}.
%
Since no inverse relation was involved when \emph{TransGaussian (SINGLE)} was trained,
to use this embedding in question answering tasks, 
we represent the inverse relations as follows:
for each relation $r$ with mean $\vdelta_r$ and variance $\mSigma_r$,
we model its inverse $r^{-1}$ as a Gaussian attention with mean $-\vdelta_r$ and variance equal to $\mSigma_r$.

We also trained TransE models on WorldCup2014 dataset by using the code released 
by the authors of \cite{guu2015traversing}.
Likewise, we use \emph{TransE (SINGLE)} to denote the model trained with 
atomic triplets only and use \emph{TransE (COMP)} to denote the model trained with the union of triplets and paths.
Note that TransE can be considered as a special case of TransGaussian where the variance matrix is the identity
and hence, the scoring formula Eq.~\eqref{eq:gaussian-conjunction} is applicable to TransE as well.

\paragraph*{Training configurations} 
For all models, dimension of entity embeddings was set to 30. 
The hidden size of LSTM was set to 80.
Word embeddings were trained jointly with the question answering model
and dimension of word embedding was set to 40.
We employed Adam \citep{kingma2014adam} as the optimizer.
All parameters were tuned on the validation set.
Under the same setting, we experimented with two cases: 
first, we trained models for path queries and conjunctive queries separately;
Furthermore, we trained a single model that addresses both types queries.
We present the results of the latter case in the next subsection
while the results of the former are included in the Appendix.

\paragraph*{Evaluation metrics}
During test time, our model receives a question in natural language and
a list of knowledge base entities contained in the question. 
Then it predicts the mean and variance of a Gaussian
attention formulated in Eq.~\eqref{eq:gaussian-conjunction} which is expected to capture the distribution of all positive answers.
We rank all entities in the knowledge base by their scores under this Gaussian attention.
Next, for each entity which is a correct answer, we check its rank relative to all incorrect answers and call this 
rank the filtered rank.  For example, if a correct entity is ranked above all negative answers except for one, it has filtered rank two. We compute this rank for all true answers
and report \emph{mean filtered rank} and \emph{H@1} which is the percentage of true answers that have filtered rank 1.

\subsection{Experimental results}
\label{sec:exp-results}

We present the results of joint learning in Table~\ref{tab:joint-res}.
These results show that TransGaussian works better than TransE in general.
In fact, \emph{TransGaussian (COMP)} achieved the best performance in almost all aspects.
Most notably, it achieved the highest H@1 rates on challenging questions such as 
``where is the club that edin dzeko plays for?'' (\#11, composition of two relations)
and ``who are the defenders on german national team?'' (\#14, conjunction of two queries).

The same table shows that TransGaussian benefits remarkably from compositional training.
For example, compositional training improved TransGaussian's H@1 rate
by near 60\% in queries on players from a given countries (\#8)
and queries on players who play a particular position (\#9).
It also boosted TransGaussian's performance on all conjunctive quries (\#13--\#15) significantly.

To understand \emph{TransGaussian (COMP)}'s weak performance on answering queries on
the professional football club located in a given country (\#10) 
and queries on professional football club that has players from a particular country (\#12),
we tested its capability of modeling the composed relation by feeding the correct relations and subjects
during test time.  It turns out that these two relations were not 
modeled well by \emph{TransGaussian (COMP)} embedding,
which limits its performance in question answering.
(See Table~\ref{tab:eva-emb} in the Appendix for quantitative evaluations.)
The same limit was found in the other three embeddings as well.

Note that all the models compared in Table \ref{tab:joint-res} uses the proposed Gaussian attention model 
because TransE is the special case of TransGaussian where the variance is fixed to one. 
Thus the main differences are whether the variance is learned and whether the embedding was trained compositionally.
Finally, we refer to Table~\ref{tab:path-res} and \ref{tab:conjunction-res} in the Appendix
for experimental results of models trained on path and conjunctive queries separately.

\begin{table}[th]
\caption{Some statistics of the WorldCup2014 dataset.}
\label{tab:data-stats}
\begin{tabular}[tb]{c|c|c|c|c }
\# entity  & \# atomic relations & \# atomic triplets & \shortstack{\# path query Q\&A \\ ( train / validation / test )}
& \shortstack{ \# conjunctive query Q\&A \\ ( train / validation / test )} \\
\hline
1127 & 6 & 3977 & 5620 / 804 / 1579 & 1564 / 224 / 420
\end{tabular}
\end{table}

\begin{table}
\centering
\caption{Results of joint learning with path queries and conjunction queries on WorldCup2014. \label{tab:joint-res}}
{\tiny
\begin{tabular}[tb]{c|l|SS|SS|SS|SS}

& & \multicolumn{2}{c|}{\parbox{1.1cm}{\centering \textbf{TransE\\(SINGLE)}}}
& \multicolumn{2}{c|}{\parbox{1.1cm}{\centering \textbf{TransE (COMP)}}}
& \multicolumn{2}{c|}{\parbox{1.1cm}{\centering \textbf{TransGaussian (SINGLE)}}}
& \multicolumn{2}{c}{\parbox{1.1cm}{\centering \textbf{TransGaussian (COMP)}}}\\
\hline
\# & Sample question &  {\parbox{0.4cm}{H@1(\%)}}  & {\parbox{0.7cm}{Mean Filtered Rank}} 
         &  {\parbox{0.4cm}{H@1(\%)}}  & {\parbox{0.7cm}{Mean Filtered Rank}} 
		 &  {\parbox{0.4cm}{H@1(\%)}}  & {\parbox{0.7cm}{Mean Filtered Rank}} 
		 &  {\parbox{0.4cm}{H@1(\%)}}  & {\parbox{0.7cm}{Mean Filtered Rank}}  \\ 
\hline \hline
1 & {\tiny \parbox{4.0cm}{ which club does alan pulido play for?}} 
& 88.59 &  1.18 & 91.95 &  1.11 & 96.64 &  1.04 & {\bf 98.66} &  {\bf 1.01} \\ 
2 & {\tiny \parbox{4.0cm}{ what position does gonzalo higuain play?}} 
& {\it 100.00} & {\it 1.00} & 98.11 &  1.03 & 98.74 &  1.01 & {\it 100.00} &  {\it 1.00}\\
3 & {\tiny \parbox{4.0cm}{ how old is samuel etoo?}}
& 67.11 &  1.44 & 90.79 &  1.13 & 94.74 &  1.08 & {\bf 97.37} &  {\bf 1.04}\\ 
4 & {\tiny \parbox{4.0cm}{ what is the jersey number of mario balotelli?}}
& 45.00 &  1.89 & 83.57 &  1.22 & 97.14 &  1.03 & {\bf 99.29} &  {\bf 1.01} \\ 
5 & {\tiny \parbox{4.0cm}{ which country is thomas mueller from ?}}
& 94.40 &  1.06 & 94.40 &  1.06 & 96.80 &  1.04 & {\bf 98.40} &  {\bf 1.02}\\ 
6 & {\tiny \parbox{4.5cm}{ which country is the soccer team fc porto based in ?}}
 & {\it 98.48} &  {\it 1.02} & {\it 98.48} & {\it 1.02} & 93.94 &  1.06 & 95.45 &  1.05 \\ \hline
7 & {\tiny \parbox{4.5cm}{ who plays professionally at liverpool fc?}}
& 95.12 &  1.10 & 90.24 &  1.20 & {\bf 98.37} &  {\it 1.04}  & 96.75 & {\it 1.04} \\ 
8 & {\tiny \parbox{4.5cm}{ which player is from iran?}}
& 89.86 &  1.51 & 76.81 &  2.07 & 38.65 &  2.96 & {\bf 99.52} & {\bf 1.00} \\ 
9 & {\tiny \parbox{4.5cm}{ name a player who plays goalkeeper?}}
& 98.96 &  1.01 & 69.79 &  1.82 & 42.71 &  5.52 & {\bf 100.00} & {\bf 1.00} \\
10 & {\tiny \parbox{4.5cm}{ which soccer club is based in mexico?}}
& 22.03 &  13.94 & {\bf 30.51} & {\bf 8.84} & 6.78 &  10.66 & 16.95 &  21.14  \\ \hline
11 & {\tiny \parbox{4.5cm}{ where is the club that edin dzeko plays for ?}} 
& 52.63 &  3.88 & 57.24 &  2.10 & 47.37 &  2.27 & {\bf 78.29} & {\bf 1.41} \\ 
12 & {\tiny \parbox{4.5cm}{ name a soccer club that has a player from australia ?}}
& 30.43 &  12.08 & {\bf 33.70} & {\bf 11.47} & 13.04 &  11.64 & 19.57 &  17.57  \\
\hline \hline
\multicolumn{2}{c|}{Overall (Path Query)} & 74.16 & 3.11 & 77.39 & {\bf 2.56} & 69.54 & 3.02 & {\bf 85.94} & 3.52\\ 
\hline \hline
13 & \parbox{4.5cm}{ {\tiny who plays forward for fc barcelona?}}
& 97.55 &  1.06 & 76.07 &  1.66 & 93.25 &  1.24 & {\bf 98.77} & {\bf 1.02} \\ 
14 & \parbox{4.5cm}{ {\tiny who are the defenders on german national team?}}
& 95.93 &  1.06 & 69.92 &  2.33 & 65.04 &  2.04 & {\bf 100.00} & {\bf 1.00} \\ 
15 & \parbox{4.5cm}{ {\tiny which player in ssc napoli is from argentina?}} 
& 88.81 &  1.17 & 76.12 &  1.76 & 88.81 &  1.35 & {\bf 97.76} & {\bf 1.03} \\
\hline  \hline
\multicolumn{2}{c|}{Overall (Conj. Query)} & 94.29 & 1.09 & 74.29 & 1.89 & 83.57 & 1.51 & {\bf 98.81} & {\bf 1.02} \\
\end{tabular}
}
\end{table}

\section{Related work}
The work of \cite{vilnis2014word} is similar to our Gaussian attention
model. They discuss many advantages of the Gaussian embedding; for
example, it is arguably a better way of handling asymmetric relations and
entailment. However the work was presented in the word2vec
\citep{mikolov2013efficient}-style word embedding setting and the Gaussian embedding
was used to capture the diversity in the meaning of a word.
Our Gaussian attention model extends their work to a more general setting
in which any memory item can be addressed through a concept represented as a
Gaussian distribution over the memory items.

\cite{bordes2014open,bordes2015large} proposed a question-answering model that embeds both questions and their answers to a common
continuous vector space. Their method in \cite{bordes2015large} can 
combine multiple knowledge bases and even generalize to a knowledge base
that was not used 
during training. However their method is limited to the simple question
answering setting in which the answer of each question associated with a
triplet in the knowledge base. In contrast, our method can handle both
composition of relations and conjunction of conditions, which are both
naturally enabled by the proposed Gaussian attention model.

\cite{neelakantan2015compositional} proposed a method that combines
relations to deal with compositional relations for knowledge base
completion. Their key technical contribution is to use recurrent neural
networks (RNNs) to {\em encode} a chain of relations. When we restrict
ourselves to path queries, question answering can be seen as a sequence
transduction task \citep{graves2012sequence,sutskever2014sequence} in
which the input is text and the output is a series of relations. If we
use RNNs as a {\em decoder}, our model would be able to handle
non-commutative composition of relations, which the current 
weighted convolution cannot handle well. Another interesting connection
to our work is that they take the maximum of the inner-product scores
\citep[see also][]{weston2013nonlinear,neelakantan2015efficient},
which are computed along multiple paths connecting a pair of
entities. Representing a set as a collection of vectors and taking the
maximum over the inner-product scores is a natural way to represent a set
of memory items. The Gaussian attention
model we propose in this paper, however, has the advantage of
differentiability and composability.

\section{Conclusion}
In this paper, we have proposed the Gaussian attention model which can be
used in a variety of contexts where we can assume that the distance
between the memory items in the latent space is compatible with some
notion of semantics. We have shown that the proposed Gaussian
scoring function can be used for knowledge base embedding achieving
competitive accuracy. We have also shown that our embedding model can
naturally propagate uncertainty when we compose relations together. Our
embedding model also benefits from compositional training proposed 
by \cite{guu2015traversing}. Furthermore, we have demonstrated the power
of the Gaussian attention model in a challenging question answering
problem which involves both composition of relations and conjunction of
queries. Future work includes experiments on natural question answering
datasets and end-to-end training including the entity extractor.

\subsection*{Acknowledgments}
The authors would like to thank Daniel Tarlow, Nate Kushman, and Kevin
Gimpel for valuable discussions.

\bibliography{iclr2017_conference}
\bibliographystyle{iclr2017_conference}

\newpage

\appendix
\section{Wordcup2014 Dataset}

\begin{center}
\begin{table}[h]
\centering
\caption{Sample atomic triplets.}
{\scriptsize
\begin{tabular}[h]{ccc}
Subject & Relation & Object \\
\hline
david\_villa & \texttt{plays\_for\_country} & spain \\
lionel\_messi & \texttt{plays\_in\_club} & fc\_barcelona\\
antoine\_griezmann & \texttt{plays\_position} & forward\\
cristiano\_ronaldo & \texttt{wears\_number} & 7 \\
fulham\_fc & \texttt{is\_in\_country} & england \\
lukas\_podolski & \texttt{is\_aged} & 29 
\end{tabular}
}
\end{table}
\end{center}

\begin{table}[h]
\caption{Statistics of the WorldCup2014 dataset.}
\centering
{\scriptsize
\begin{tabular}{r|c}
\hline
\# entity & 1127 \\
\# atomic relations & 6 \\
\# atomic triplets & 3977 \\
\hline
\# relations (atomic and compositional) in path queries & 12 \\
\# question and answer pairs in path queries ( train / validation / test ) & 5620 / 804 / 1579 \\
\hline  
\# types of questions in conjunctive queries & 3 \\
\# question and answer pairs in conjunctive queries ( train / validation / test ) & 1564 / 224 / 420 \\
\hline
size of vocabulary & 1781 \\
\hline 
\end{tabular}
}
\end{table}

\begin{table}[!h]
\centering
\caption{Templates of questions. In the table, (player), (club), (position) are placeholders
of named entities with associated type. (country\_1) is a placeholder for a country name while 
(country\_2) is a placeholder for the adjectival form of a country.
\label{tab:query-templates}}
\begin{tiny}
\begin{tabular}{c|l|l}
\hline
\# & Query template & Question template \\
\hline 
\multirow{3}{*}{1} &
\multirow{3}{*}{Find $e\in\mathcal{E}$: ( \playerblank{}, \texttt{plays\_in\_club},$e$) is true }
& which club does \playerblank{} play for ?  \\ 
 & & which professional football team does \playerblank{} play for ? \\
 & & which football club does \playerblank{} play for ? \\
\hline
2 &
Find $e\in\mathcal{E}$: (\playerblank{}, \texttt{plays\_position}, $e$) is true
& what position does \playerblank{} play ? \\
\hline
\multirow{2}{*}{3} &
\multirow{2}{*}{Find $e\in\mathcal{E}$: (\playerblank{}, \texttt{is\_aged}, $e$) is true} 
& how old is \playerblank{} ?  \\
 & & what is the age of \playerblank{} ? \\
\hline
\multirow{2}{*}{4} &
\multirow{2}{*}{Find $e\in\mathcal{E}$: (\playerblank{}, \texttt{wears\_number}, $e$) is true}  
& what is the jersey number of \playerblank{} ? \\
 & & what number does \playerblank{} wear ? \\
\hline
\multirow{3}{*}{5} &
\multirow{3}{*}{Find  $e\in\mathcal{E}$: (\playerblank{}, \texttt{plays\_for\_country}, $e$) is true} 
& what is the nationality of \playerblank{} ?  \\
 & & which national team does \playerblank{} play for ? \\
 & & which country is \playerblank{} from ? \\
\hline
6 &
Find $e\in\mathcal{E}$: (\clubblank{}, \texttt{is\_in\_country}, $e$) is true
& which country is the soccer team \clubblank{} based in ? \\
\hline
\multirow{4}{*}{7} &
\multirow{4}{*}{Find $e\in\mathcal{E}$: (\clubblank{}, $\texttt{plays\_in\_club}^{-1}$, $e$) is true} 
 & name a player from \clubblank{} ?  \\
 & & who plays at the soccer club \clubblank{} ? \\ 
 & & who is from the professional football team \clubblank{} ? \\
 & & who plays professionally at \clubblank{} ? \\
\hline
\multirow{4}{*}{8} &
\multirow{4}{*}{Find $e\in\mathcal{E}$: (\countryblanka{}, $\texttt{plays\_for\_country}^{-1}$, $e$) is true} 
& which player is from \countryblanka{} ? \\
 & & name a player from \countryblanka{} ? \\
 & & who is from \countryblanka{} ? \\ 
 & & who plays for the \countryblanka{} national football team ? \\
\hline
\multirow{2}{*}{9} &
\multirow{2}{*}{Find $e\in\mathcal{E}$: (\positionblank{}, $\texttt{plays\_position}^{-1}$, $e$) is true} 
& name a player who plays \positionblank{} ? \\
 & & who plays \positionblank{} ? \\
\hline
\multirow{2}{*}{10} &
\multirow{2}{*}{Find $e\in\mathcal{E}$: (\countryblanka{}, $\texttt{is\_in\_country}^{-1}$, $e$) is true} 
& which soccer club is based in \countryblanka{} ? \\
 & & name a soccer club in \countryblanka{} ? \\
\hline
\multirow{2}{*}{11} &
\multirow{2}{*}{Find $e\in\mathcal{E}$: (\playerblank{}, \texttt{plays\_in\_club / is\_in\_country}, $e$) is true} 
& which country does \playerblank{} play professionally in ? \\
 & & where is the football club that \playerblank{} plays for ? \\
\hline
\multirow{3}{*}{12} &
\multirow{3}{*}{Find $e\in\mathcal{E}$: (\countryblanka{}, $\texttt{plays\_for\_country}^{-1}$ / \texttt{plays\_in\_club}, $e$) is true} 
& which professional football team do players from \countryblanka{} play for ? \\
 & & name a soccer club that has a player from \countryblanka{} ? \\ 
 & & which professional football team has a player from \countryblanka{} ? \\
\hline
13 & \parbox{6.0cm}{
Find $e\in\mathcal{E}$: (\positionblank{}, $\texttt{plays\_position}^{-1}$, $e$) is true and
(\clubblank{}, $\texttt{plays\_in\_club}^{-1}$, $e$) is true }
& \parbox{5.0cm}{who plays \positionblank{} for \clubblank{}? \\ 
who are the \positionblank{} at \clubblank{} ? \\
name a \positionblank{} that plays for \clubblank{} ?} \\
\hline
14 & \parbox{6.0cm}{
Find $e\in\mathcal{E}$: (\positionblank{}, $\texttt{plays\_position}^{-1}$, $e$) is true and
(\countryblanka{}, $\texttt{plays\_for\_country}^{-1}$, $e$) is true }
& \parbox{5.0cm}{who plays \positionblank{} for \countryblanka{} ?\\ 
who are the \positionblank{} on \countryblanka{} national team ? \\
name a \positionblank{} from \countryblanka{} ? \\
which \countryblankb{} footballer plays \positionblank{} ? \\
name a \countryblankb{} \positionblank{} ? }
\\ \hline
15  & \parbox{6.0cm}{
Find $e\in\mathcal{E}$: (\clubblank{}, $\texttt{plays\_in\_club}^{-1}$, $e$) is true and
(\countryblanka{}, $\texttt{plays\_for\_country}^{-1}$, $e$) is true}
& \parbox{5.0cm}{who are the \countryblankb{} players at \clubblank{} ? 
\\ 
which \countryblankb{} footballer plays for \clubblank{} ? \\
name a \countryblankb{} player at \clubblank{} ? \\
which player in \clubblank{} is from \countryblanka{} ?}\\ \hline
\end{tabular}
\end{tiny}
\end{table}

\begin{table}[!h]
\centering
\caption{(Composed) relations and sample questions in path queries. \label{tab:path-query-samples}}
\begin{tiny}
\begin{tabular}{c|c|c|l|l}
\hline
\# & Relation & Type & Sample question & Sample answer \\
\hline 
\multirow{2}{*}{1} & \multirow{2}{*}{\tt{plays\_in\_club}} & \multirow{2}{*}{many-to-one }
& which club does alan pulido play for ? & tigres\_uanl  \\ 
& & & which professional football team does klaas jan huntelaar play for ? & fc\_schalke\_04 \\ \hline
2 & \tt{plays\_position} & many-to-one & what position does gonzalo higuain play ? & ssc\_napoli \\ \hline
\multirow{2}{*}{3} & \multirow{2}{*}{\tt{is\_aged}} & \multirow{2}{*}{many-to-one}
& how old is samuel etoo ? & 33 \\
& & & what is the age of luis suarez ? & 27\\ \hline
\multirow{2}{*}{4} & \multirow{2}{*}{\tt{wears\_number}} & \multirow{2}{*}{many-to-one} 
& what is the jersey number of mario balotelli ? & 9 \\
& & & what number does shinji okazaki wear ? & 9 \\ \hline
\multirow{2}{*}{5} & \multirow{2}{*}{\tt{plays\_for\_country}} & \multirow{2}{*}{many-to-one}
& which country is thomas mueller from ? & germany \\
& & & what is the nationality of helder postiga ? & portugal \\ \hline
6 & \tt{is\_in\_country} & many-to-one & which country is the soccer team fc porto based in ? & portugal \\
\hline
\multirow{2}{*}{7} & \multirow{2}{*}{$\texttt{plays\_in\_club}^{-1}$} & \multirow{2}{*}{one-to-many}
 & who plays professionally at liverpool fc ? & steven\_gerrard \\
 & & & name a player from as roma ? & miralem\_pjanic \\ \hline
\multirow{2}{*}{8} & \multirow{2}{*}{$\texttt{plays\_for\_country}^{-1}$} & \multirow{2}{*}{one-to-many}
 & which player is from iran ? & masoud\_shojaei \\
 & & & name a player from italy ? & daniele\_de\_rossi \\ \hline
\multirow{2}{*}{9} & \multirow{2}{*}{$\texttt{plays\_position}^{-1}$} & \multirow{2}{*}{one-to-many}
 & name a player who plays goalkeeper ? & gianluiqi\_buffon \\
 & & & who plays forward ? & raul\_jimenez \\ \hline
\multirow{2}{*}{10} & \multirow{2}{*}{$\texttt{is\_in\_country}^{-1}$} & \multirow{2}{*}{one-to-many} & which soccer club is based in mexico ? & cruz\_azul\_fc \\
& & & name a soccer club in australia ? & melbourne\_victory\_fc \\ \hline
\multirow{2}{*}{11} & \multirow{2}{*}{\tt{plays\_in\_club / is\_in\_country }} & \multirow{2}{*}{many-to-one}
 & where is the club that edin dzeko plays for ? & england \\
 & & & which country does sime vrsaljko play professionally in ? & italy \\ \hline
\multirow{2}{*}{12} & \multirow{2}{*}{$\texttt{plays\_for\_country}^{-1}$ / \tt{plays\_in\_club}} & \multirow{2}{*}{many-to-many}
 & name a soccer club that has a player from australia ? & crystal\_palace\_fc \\
 & & & name a soccer club that has a player from spain ? & fc\_barcelona \\ 
\hline
\end{tabular}
\end{tiny}
\end{table}

\begin{table}[!h]
\centering
\caption{Conjunctive queries and sample questions. \label{tab:conjunction-query-samples}}
\begin{tiny}
\begin{tabular}{c|c|c|l|l}
\hline
\# & Relations & Sample questions & Entities in questions & Sample answer \\
\hline 
13 & \parbox{3.0cm}{\centering  $\texttt{plays\_position}^{-1}$ \\ and  \\ $\texttt{plays\_in\_club}^{-1}$ }
& \parbox{5.0cm}{who plays forward for fc barcelona ? \\ who are the midfielders at fc bayern muenchen ?} 
& \parbox{3.0cm}{forward ,  fc\_barcelona \\ midfielder, fc\_bayern\_muenchen} 
& \parbox{3.0cm}{lionel\_messi  \\ toni\_kroos }\\
\hline
14 & \parbox{3.0cm}{\centering  $\texttt{plays\_position}^{-1}$ \\ and \\ $\texttt{plays\_for\_country}^{-1}$ }
& \parbox{5.0cm}{who are the defenders on german national team ? \\ which mexican footballer plays forward ? }
& \parbox{3.0cm}{defender , germany \\ defender , mexico}
& \parbox{3.0cm}{ per\_mertesacker \\ raul\_jimenez} \\
\hline
15 & \parbox{3.0cm}{\centering  $\texttt{plays\_in\_club}^{-1}$ \\ and \\ $\texttt{plays\_for\_country}^{-1}$ }
& \parbox{5.0cm}{which player in paris saint-germain fc is from argentina ? \\ who are the korean players at beijing guoan ?}
& \parbox{3.0cm}{paris\_saint-germain\_fc , argentina \\ beijing\_guoan , korea }
& \parbox{3.0cm}{ezequiel\_lavezzi \\ ha\_daesung} \\
\hline
\end{tabular}
\end{tiny}
\end{table}

\section{TransGaussian embedding of WorldCup2014}

\begin{figure}[tb]
 \begin{center}
  \includegraphics[width=\textwidth]{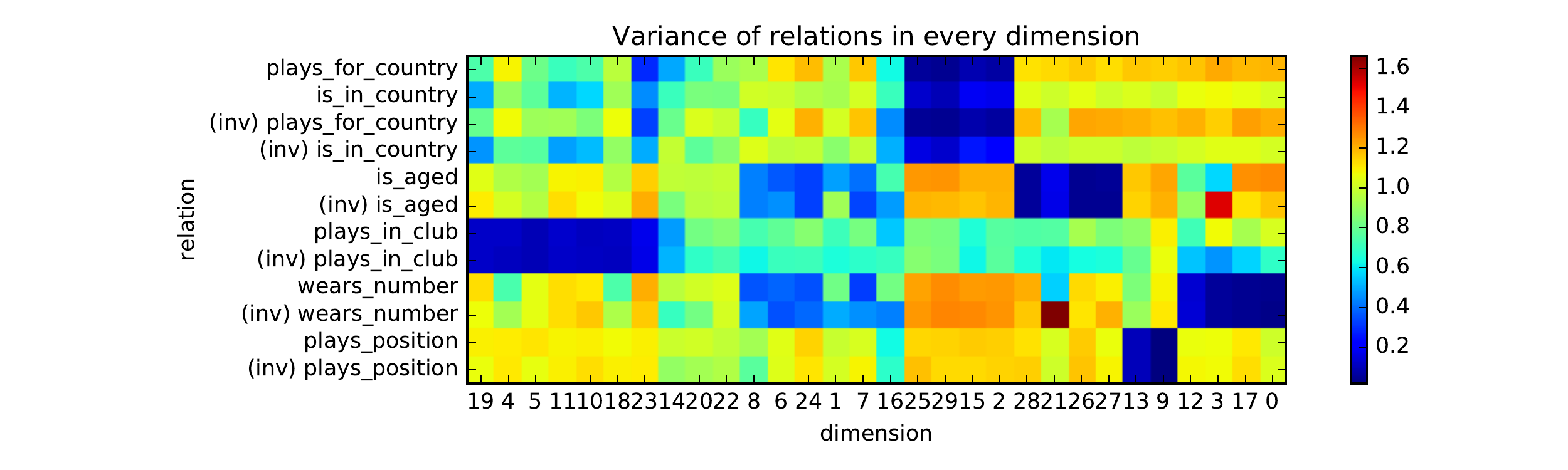}
 \caption{Variance of each relation.  Each row shows the diagonal values in the variance matrix associated with 
 a relation.  Columns are permuted to reveal the block structure.
 \label{fig:transGaussian-var}}
 \end{center}
\end{figure}

\begin{figure}[tb]
\begin{center}
   \begin{subfigure}[t]{0.4\textwidth}
   \includegraphics[height=5cm]{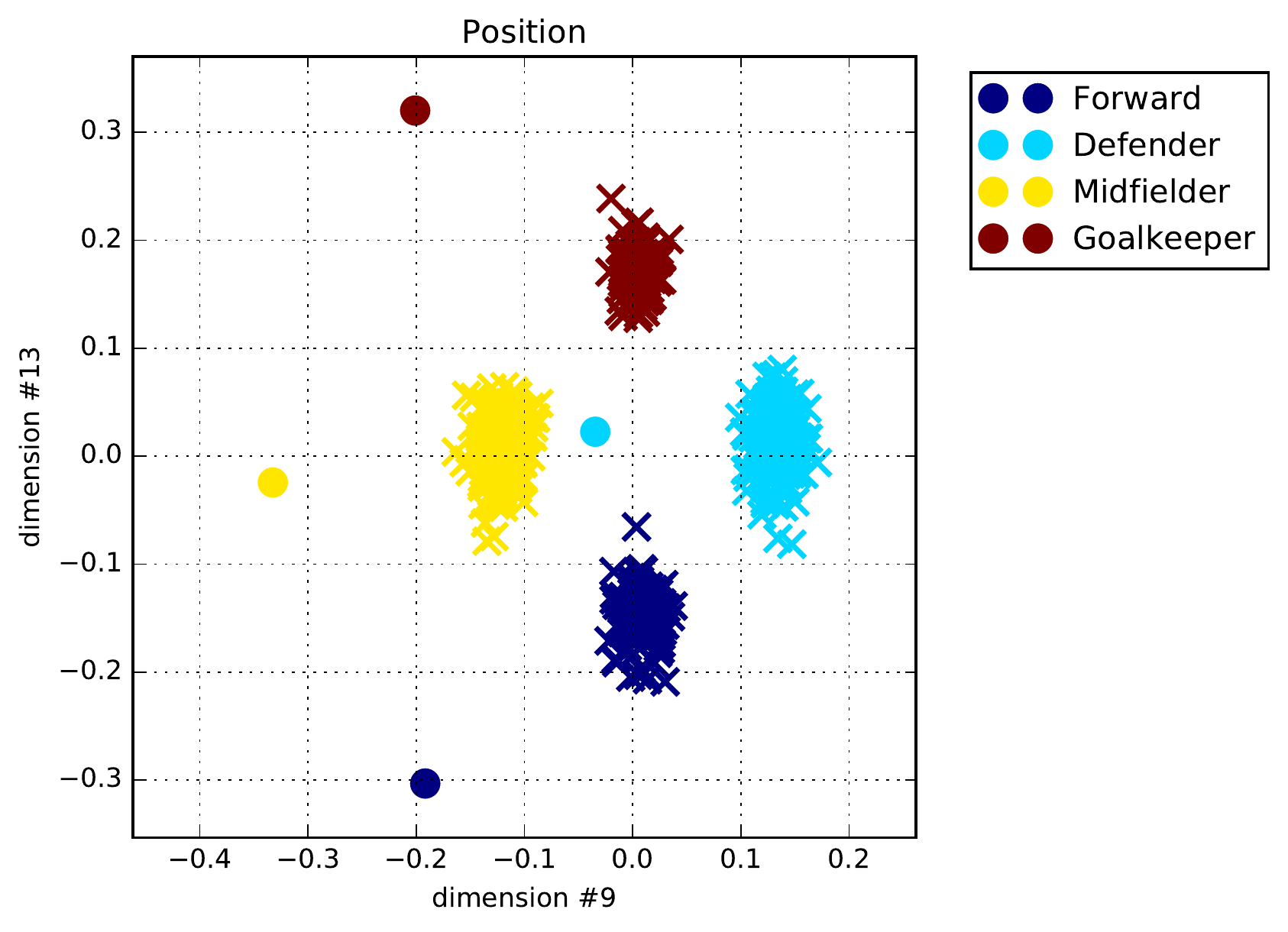}
   \caption{\texttt{plays\_position}}
  \end{subfigure}
  \hfill
   \begin{subfigure}[t]{0.4\textwidth}
   \includegraphics[height=5cm]{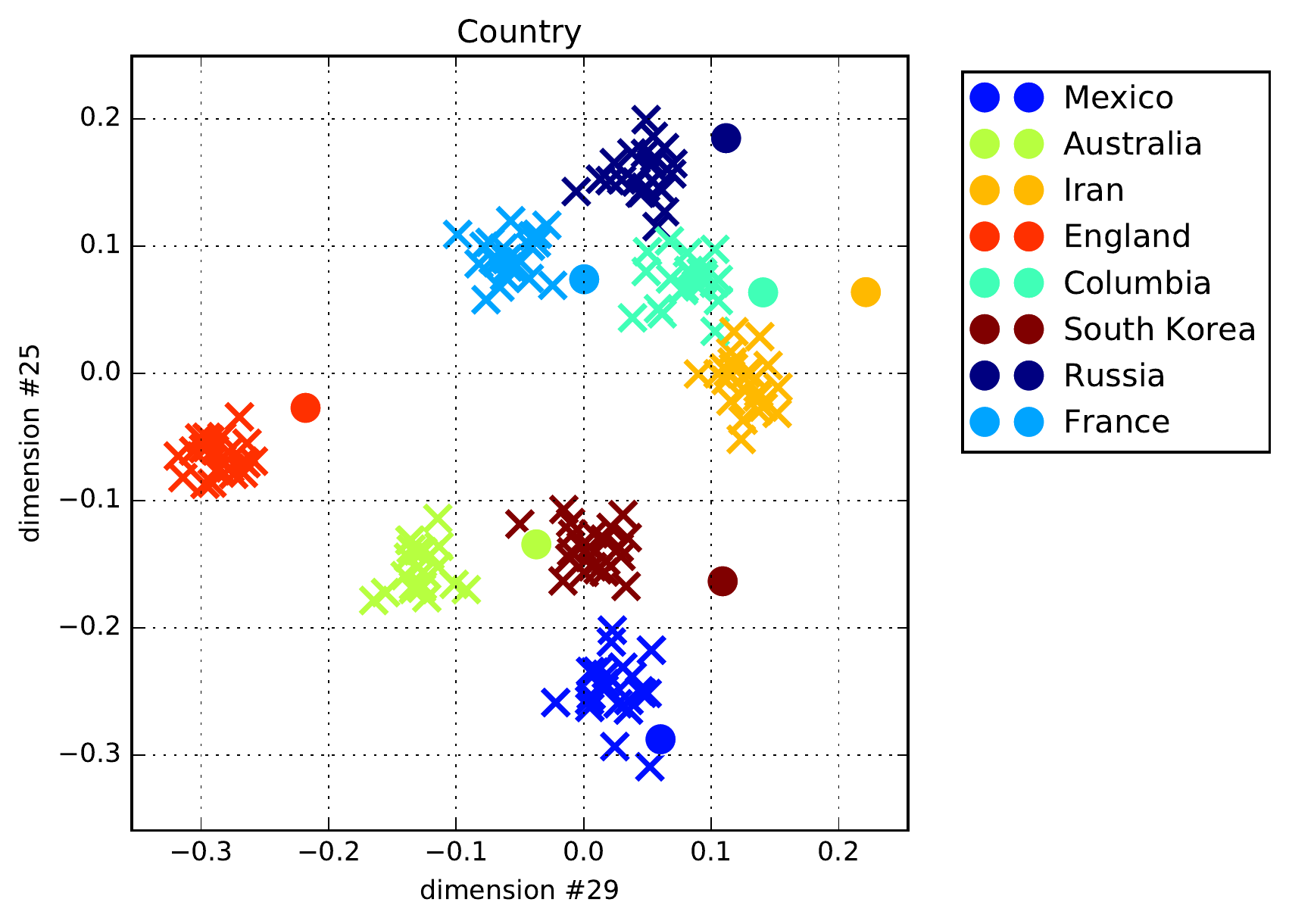}
   \caption{\texttt{plays\_for\_country}}
  \end{subfigure}
  \\
   \begin{subfigure}[t]{0.4\textwidth}
   \includegraphics[height=5cm]{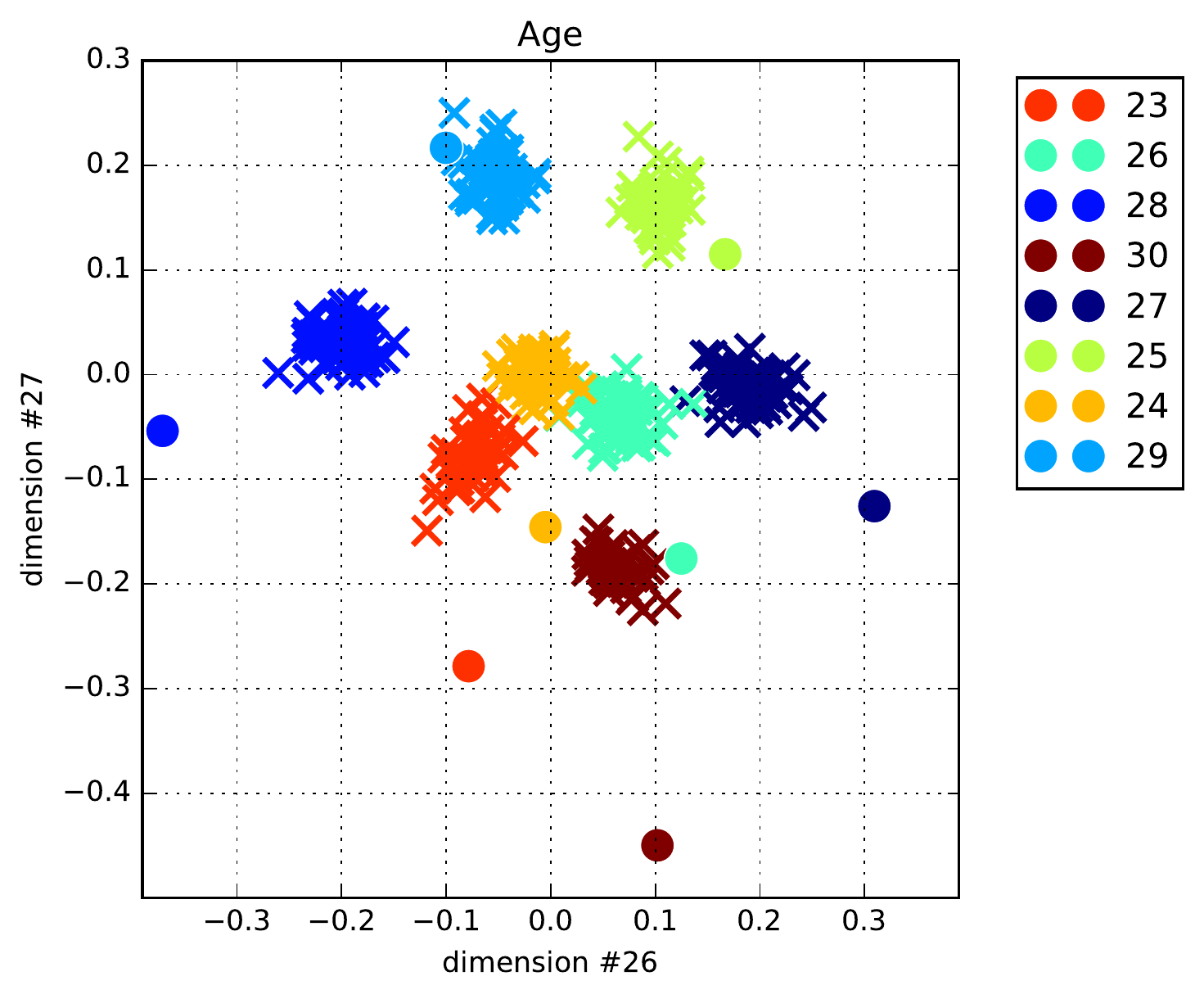}
   \caption{\texttt{is\_aged}}
  \end{subfigure}
  \hfill
   \begin{subfigure}[t]{0.4\textwidth}
   \includegraphics[height=5cm]{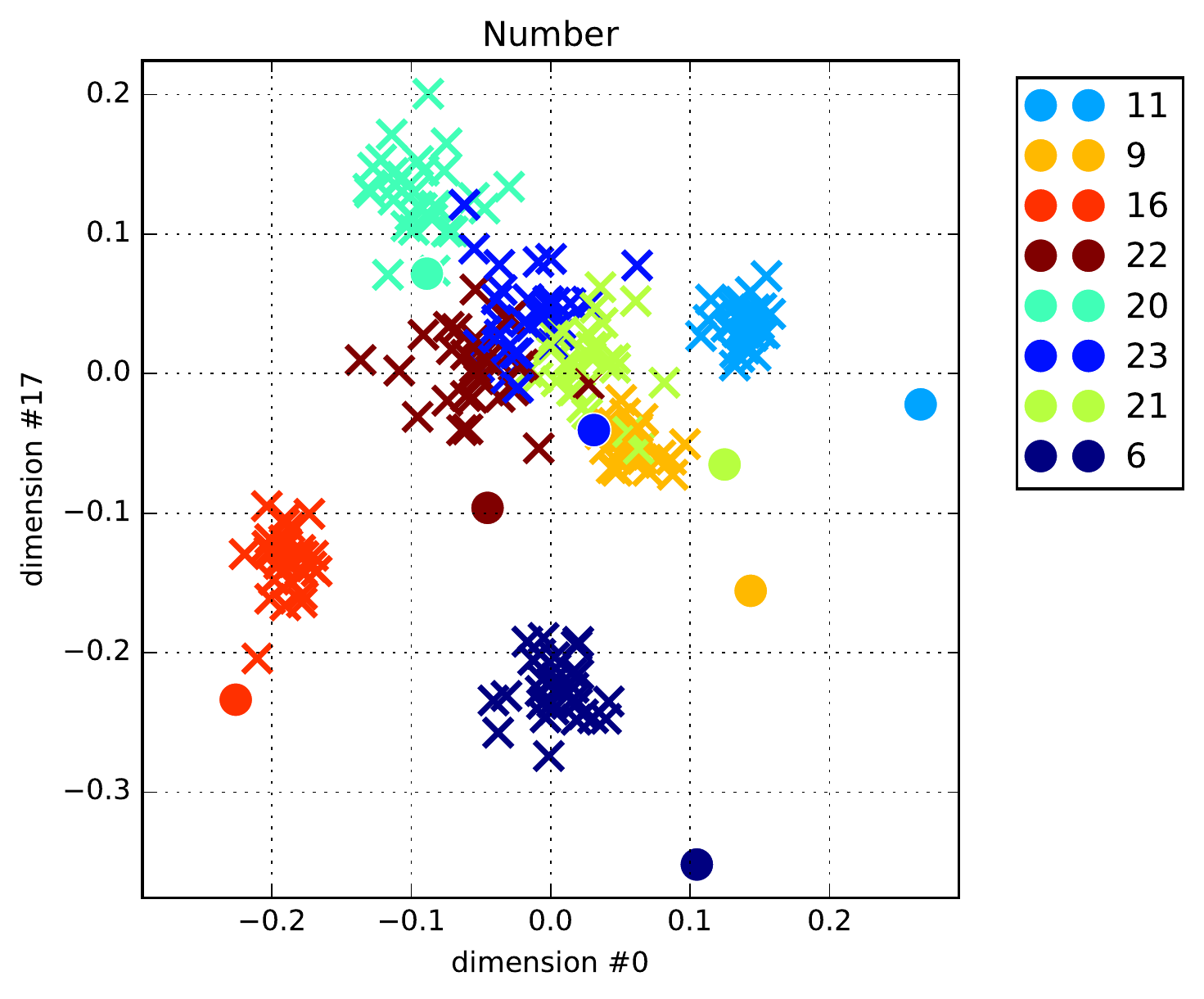}
   \caption{\texttt{wears\_number}}
  \end{subfigure}
  \\
   \begin{subfigure}[t]{0.4\textwidth}
   \includegraphics[height=5cm]{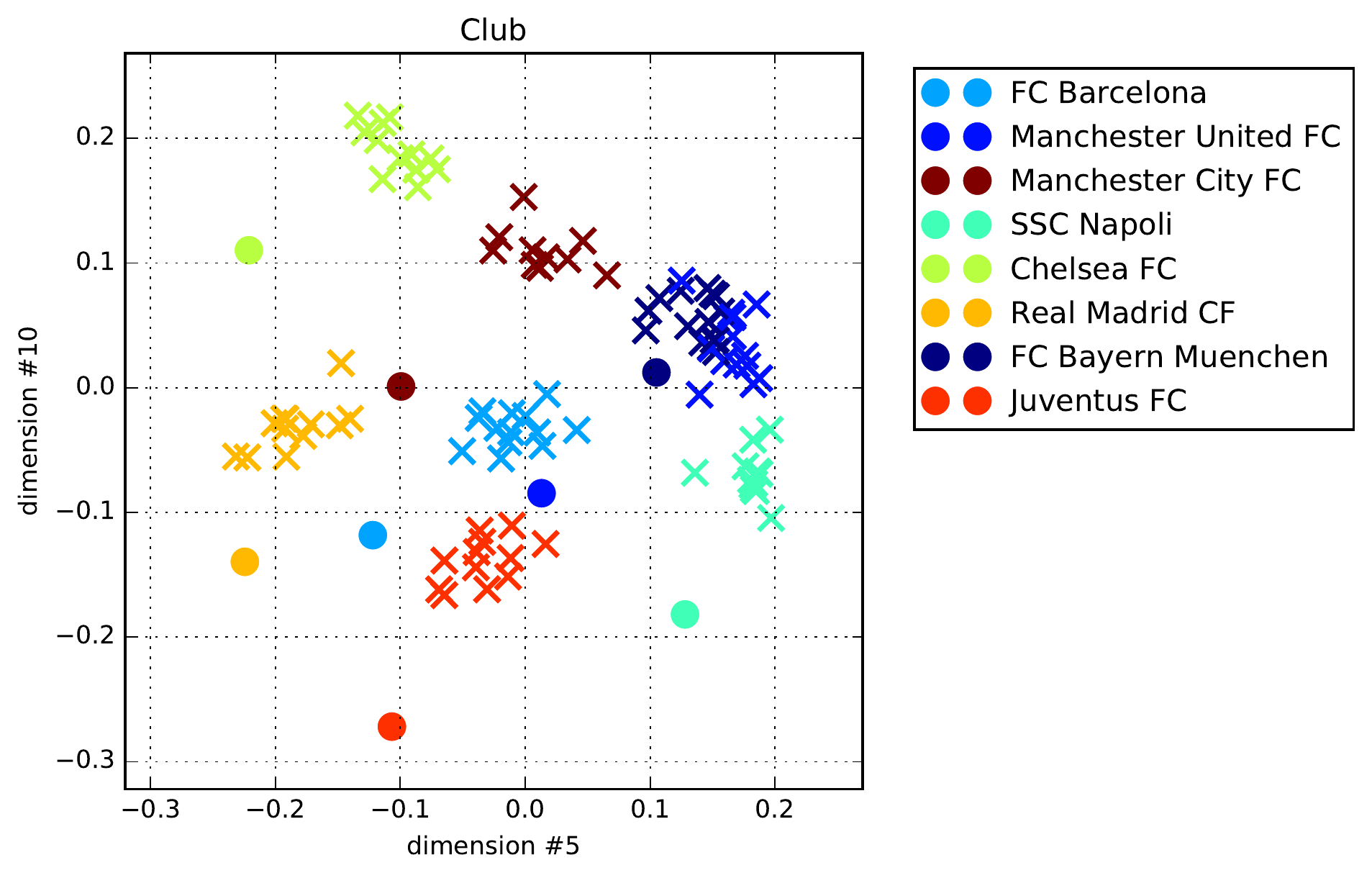}
   \caption{\texttt{plays\_in\_club}}
  \end{subfigure}
  \hfill
   \begin{subfigure}[t]{0.4\textwidth}
   \includegraphics[height=5cm]{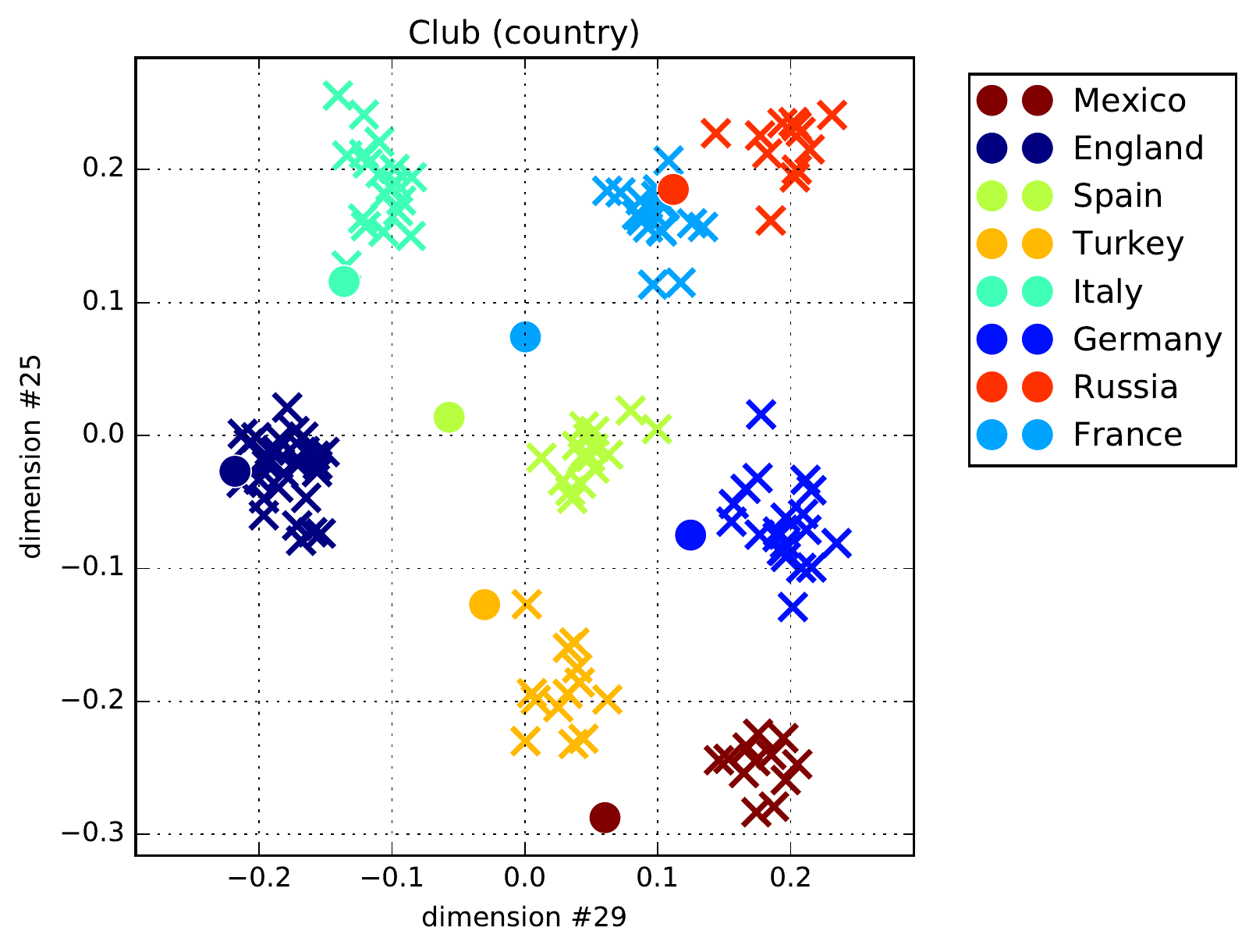}
   \caption{\texttt{is\_in\_country}}
  \end{subfigure} 
  \\        
 \caption{TransGaussian entity embeddings.
 Crosses are the subjects and 
 circles are the objects of a relation.
 Specifically, crosses are players in (a)-(e)
 and professional football clubs in (f).
 \label{fig:tranGaussian-emb}}
 \end{center}
\end{figure}

We trained our TransGaussian model on triplets and paths from WorldCup2014 dataset and illustrated the 
embeddings in Fig~\ref{fig:transGaussian-var} and \ref{fig:tranGaussian-emb}.
Recall that we modeled every relation as a Gaussian with diagonal
covariance matrix.  Fig~\ref{fig:transGaussian-var} shows the learned
variance parameters of different relations.
Each row corresponds to the variances of one relation.
Columns are permuted to reveal the block structure.
From this figure, we can see that every relation has a small variance in two or more dimensions.
This implies that the coordinates of the embedding space are partitioned
into semantically coherent clusters each of which represent 
a particular attribute of a player (or a football club).
To verify this further, we picked the two coordinates in which a relation (e.g. \texttt{plays\_position})
has the least variance and projected the embedding of all valid subjects and objects (e.g. players and positions)
of the relation to this 2 dimensional subspace. See
Fig.~\ref{fig:tranGaussian-emb}.
The relation between the subjects and the objects are simply translation
in the projection when the corresponding subspace is two dimensional (e.g.,
\texttt{plays\_position} relation in Fig.~\ref{fig:tranGaussian-emb}
(a)). The same is true for other relations that requires larger
dimension but it is more challenging to visualize in two dimensions.
For relations that have a large number of unique objects, we only plotted for the eight objects with the most subjects
for clarity of illustration.

Furthermore, in order to elucidate whether we are limited by the
capacity of the TransGaussian embedding or the ability to decode
question expressed in natural language, we evaluated the test
question-answer pairs using the TransGaussian embedding composed
according to the ground-truth relations and entities.
The results were evaluated with the same metrics as in Sec.~\ref{sec:exp-results}.  
This estimation is conducted for TransE embeddings as well.  See Table~\ref{tab:eva-emb} for the results.
Compared to Table \ref{tab:joint-res}, the accuracy of TransGaussian
(COMP) is higher on the atomic relations and path queries but lower on
conjunctive queries. This is natural because when the query is
simple there is not much room for the question-answering network to
improve upon just combining the relations according to the ground truth
relations, whereas when the query is complex the network could combine
the embedding in a more creative way to overcome its limitation.
In fact, the two queries (\#10 and \#12) that TransGaussian (COMP) did
not perform well in Table \ref{tab:joint-res} pertain to a single relation
$\texttt{is\_in\_country}^{-1}$ (\#10) and a composition of two
relations $\texttt{plays\_for\_country}^{-1}$ / \texttt{plays\_in\_club}
(\#12). The performance of the two queries were low even when the ground
truth relations were given, which indicates that the
TransGaussian embedding rather than the question-answering network is
the limiting factor.

%
%

\begin{table}
\centering
\caption{Evaluation of embeddings. 
We evaluate the embeddings by feeding the correct entities and relations from a path or conjunctive query to an embedding model and using its scoring function to retrieve the answers from the embedded knowledge base.
\label{tab:eva-emb}}
{\scriptsize
\begin{tabular}[tb]{c|c|SS|SS|SS|SS}

& & \multicolumn{2}{c|}{\parbox{1.5cm}{\centering \textbf{TransE\\(SINGLE)}}}
& \multicolumn{2}{c|}{\parbox{1.5cm}{\centering \textbf{TransE (COMP)}}}
& \multicolumn{2}{c|}{\parbox{1.5cm}{\centering \textbf{TransGaussian (SINGLE)}}}
& \multicolumn{2}{c}{\parbox{1.5cm}{\centering \textbf{TransGaussian (COMP)}}}\\
\hline
\# & Relation &  {\parbox{0.7cm}{H@1(\%)}}  & {\parbox{0.8cm}{Mean Filtered Rank}} 
         &  {\parbox{0.7cm}{H@1(\%)}}  & {\parbox{0.8cm}{Mean Filtered Rank}} 
		 &  {\parbox{0.7cm}{H@1(\%)}}  & {\parbox{0.8cm}{Mean Filtered Rank}} 
		 &  {\parbox{0.7cm}{H@1(\%)}}  & {\parbox{0.8cm}{Mean Filtered Rank}}  \\ 
\hline \hline
1 & {\tiny \tt{plays\_in\_club}} & 75.54  &  1.38 & 93.48 & 1.09 & {\bf 99.86} & {\bf 1.00} & 98.51 & 1.02 \\
2 & {\tiny \tt{plays\_position}} & 96.33  &  1.04 & 94.02 & 1.09 & 98.37 & 1.02 & {\bf 100.00} & {\bf 1.00} \\
3 & {\tiny \tt{is\_aged}} & 55.03  &  1.69 & 91.44 & 1.12 & 96.88 & 1.03 &{\bf  100.00} & {\bf 1.00} \\
4 & {\tiny \tt{wears\_number}} & 38.86  &  2.09 & 78.67 & 1.32 & 95.92 & 1.04 & {\bf 100.00} & {\bf 1.00} \\
5 & {\tiny \tt{plays\_for\_country}} & 71.60  &  1.39 & 94.84 & 1.10 & 99.32 & 1.01 & {\bf 100.00} & {\bf 1.00} \\
6 & {\tiny \tt{is\_in\_country}} & 98.32  &  1.03 & 99.66 & 1.00 & 99.33 & 1.01 & {\bf 100.00} & {\bf 1.00} \\
\hline
7 & {\tiny $\texttt{plays\_in\_club}^{-1}$} & 87.50  &  1.46 & 83.42 & 1.45 & 94.70 & 1.07 & {\bf 97.42} & {\bf 1.03} \\
8 & {\tiny $\texttt{plays\_for\_country}^{-1}$} & 82.47  &  1.68 & 68.21 & 3.37 & 25.27 & 5.66 & {\bf 98.78} & {\bf 1.02} \\
9 & {\tiny $\texttt{plays\_position}^{-1}$} & {\bf 100.00}  &  {\bf 1.00} & 75.54 & 1.60 & 13.59 & 24.35 & 98.78 & 1.02 \\
10 & {\tiny $\texttt{is\_in\_country}^{-1}$} & 23.11  &  26.92 & {\bf 23.48} & {\bf 23.27} & 8.32 & 130.59 & 19.41 & 83.61 \\
\hline
11 & {\tiny \tt{plays\_in\_club / is\_in\_country }} 
& 20.24  &  7.05 & 58.29 & 1.98 & 46.88 & 2.99 & {\bf 80.16} & {\bf 1.38} \\
12 & {\tiny $\texttt{plays\_for\_country}^{-1}$ / \tt{plays\_in\_club}} 
& 25.32  &  22.27 & {\bf 27.73} & {\bf 10.04} &  19.04 & 35.59 & 20.15 & 33.01 \\
\hline \hline
\multicolumn{2}{c|}{Overall (Path relations)} & 64.64 & 5.09 & 75.02 & {\bf 3.59} & 67.22 & 14.87 & {\bf 86.73} & 8.79 \\ 
\hline \hline
13 & \parbox{3.0cm}{\centering {\tiny $\texttt{plays\_position}^{-1}$ \\ and \\ $\texttt{plays\_in\_club}^{-1}$}} 
& 91.85 & 1.20 & 69.97 & 1.82 & 77.45 & 1.83 & {\bf 95.38} & {\bf 1.06} \\ \hline
14 & \parbox{3.0cm}{\centering {\tiny $\texttt{plays\_position}^{-1}$\\ and \\ $\texttt{ plays\_for\_country}^{-1}$}}
& 91.71 & 1.23 & 66.71 & 2.85 & 51.49 & 4.88 & {\bf 97.83} & {\bf 1.05} \\  \hline
15 & \parbox{3.0cm}{\centering {\tiny $\texttt{ plays\_in\_club}^{-1}$ \\ and \\ $\texttt{ is\_in\_country}^{-1}$}} 
& 88.59 & 1.20 & 73.37 & 1.80 & 83.42 & 1.34 & {\bf 94.70} & {\bf 1.08} \\
\hline  \hline
\multicolumn{2}{c|}{Overall (Conj. relations)} & 90.72 & 1.21 & 70.02 & 2.16 & 70.79 & 2.68 & {\bf 95.97} & {\bf 1.06} \\
\end{tabular}
}
\end{table}

\begin{table}
\centering
\caption{Experimental results of path queries on WorldCup2014. \label{tab:path-res}}
{\tiny
\begin{tabular}[htb]{c|l|SS|SS|SS|SS}
& &  \multicolumn{2}{c|}{\parbox{1.1cm}{\centering \textbf{TransE\\(SINGLE)}}}
& \multicolumn{2}{c|}{\parbox{1.1cm}{\centering \textbf{TransE (COMP)}}}
& \multicolumn{2}{c|}{\parbox{1.1cm}{\centering \textbf{TransGaussian (SINGLE)}}}
& \multicolumn{2}{c}{\parbox{1.1cm}{\centering \textbf{TransGaussian (COMP)}}}\\
\hline
\# & Relation and sample question &  {\parbox{0.4cm}{H@1(\%)}}  & {\parbox{0.7cm}{Mean Filtered Rank}} 
         &  {\parbox{0.4cm}{H@1(\%)}}  & {\parbox{0.7cm}{Mean Filtered Rank}} 
		 &  {\parbox{0.4cm}{H@1(\%)}}  & {\parbox{0.7cm}{Mean Filtered Rank}} 
		 &  {\parbox{0.4cm}{H@1(\%)}}  & {\parbox{0.7cm}{Mean Filtered Rank}}  \\ 
\hline \hline
1 & {\tiny \parbox{4.0cm}{ \texttt{plays\_in\_club} \\ (which club does alan pulido play for?)}} 
& 90.60 &  1.12 & 92.62 &  1.11 & 96.64 &  {\it 1.03} & {\bf 97.99} &  {\it 1.03} \\ \hline
2 & {\tiny \parbox{4.0cm}{ \texttt{plays\_position} \\ (what position does gonzalo higuain play?)}} 
& {\it 100.00} &  {\it 1.00} & 98.11 &  1.02 & 98.74 &  1.01 & {\it 100.00} &  {\it 1.00}\\ \hline
3 & {\tiny \parbox{4.0cm}{ \texttt{is\_aged} \\ (how old is samuel etoo?)}}
& 81.58 &  1.30 & 92.11 &  1.10 & 96.05 &  1.04 & {\bf 100.00} &  {\bf 1.00}\\ \hline
4 & {\tiny \parbox{4.0cm}{ \texttt{wears\_number} \\ (what is the jersey number of mario balotelli?)}}
& 44.29 &  1.88 & 85.71 &  1.19 & 96.43 &  1.04 & {\bf 100.00} &  {\bf 1.00} \\ \hline
5 & {\tiny \parbox{4.0cm}{ \texttt{plays\_for\_country} \\ (which country is thomas mueller from ?)}}
& 97.60 &  1.02 & 94.40 &  1.11 & 98.40 &  1.02 & {\bf 99.20} &  {\bf 1.01}\\ \hline
6 & {\tiny \parbox{4.7cm}{ \texttt{is\_in\_country} \\ (which country is the soccer team fc porto based in ?)}}
& {\it 98.48} &  {\it 1.02} & {\it 98.48} &  {\it 1.02} & 93.94 &  1.08 & {\it 98.48} &  {\it 1.02} \\ \hline
7 & {\tiny \parbox{4.7cm}{ $\texttt{plays\_in\_club}^{-1}$ \\ (who plays professionally at liverpool fc?)}}
& 95.12 &  1.08 & 86.99 &  1.38 & {\it 96.75} &  {\it 1.03} & {\it 96.75} &  {\it 1.03} \\ \hline
8 & {\tiny \parbox{4.7cm}{ $\texttt{plays\_for\_country}^{-1}$ \\ (which player is from iran?)}}
& 81.16 &  1.61 & 72.46 &  2.36 & 40.58 &  3.19 & {\bf 93.24} &  {\bf 1.48} \\ \hline
9 & {\tiny \parbox{4.7cm}{ $\texttt{plays\_position}^{-1}$ \\ (name a player who plays goalkeeper?)}}
& {\bf 100.00} &  {\bf 1.00} & 30.21 &  2.30 & 55.21 &  5.09 & 85.42 &  1.15 \\ \hline
10 & {\tiny \parbox{4.7cm}{ $\texttt{is\_in\_country}^{-1}$ \\ (which soccer club is based in mexico?)}}
& {\bf 24.58} &  11.47 & 23.73 &  10.07 & 5.08 &  {\bf 9.18} & 17.80 &  20.10  \\ \hline
11 & {\tiny \parbox{4.7cm}{ \texttt{plays\_in\_club / is\_in\_country } \\ (where is the club that edin dzeko plays for ?)}}
& 48.68 &  4.24 & 62.50 &  2.07 & 48.03 &  2.41 & {\bf 76.97} &  {\bf 1.50} \\ \hline
12 & {\tiny \parbox{4.7cm}{ $\texttt{plays\_for\_country}^{-1}$ / \texttt{plays\_in\_club} \\
(name a soccer club that has a player from australia ?)}} & {\bf 34.78} &  {\bf 9.49} & 30.43 &  11.26 & 6.52 &  9.88 & 16.30 &  20.27  \\
\hline  \hline
\multicolumn{2}{c|}{Overall} & 74.92 & 2.80 & 74.35 & {\bf 2.71} & 70.17 & 2.82 & {\bf 84.42} & 3.68 
\end{tabular}
}
\end{table}

\begin{table}[htb]
\centering
\caption{Experimental results of conjunctive queries on WorldCup2014. \label{tab:conjunction-res}}
{\tiny
\begin{tabular}[tb]{c|l|SS|SS|SS|SS}
& & \multicolumn{2}{c|}{\parbox{1.1cm}{\centering \textbf{TransE\\(SINGLE)}}}
& \multicolumn{2}{c|}{\parbox{1.1cm}{\centering \textbf{TransE (COMP)}}}
& \multicolumn{2}{c|}{\parbox{1.1cm}{\centering \textbf{TransGaussian (SINGLE)}}}
& \multicolumn{2}{c}{\parbox{1.1cm}{\centering \textbf{TransGaussian (COMP)}}}\\
\hline
\# & Relation and sample question &  {\parbox{0.4cm}{H@1(\%)}}  & {\parbox{0.7cm}{Mean Filtered Rank}} 
         &  {\parbox{0.4cm}{H@1(\%)}}  & {\parbox{0.7cm}{Mean Filtered Rank}} 
		 &  {\parbox{0.4cm}{H@1(\%)}}  & {\parbox{0.7cm}{Mean Filtered Rank}} 
		 &  {\parbox{0.4cm}{H@1(\%)}}  & {\parbox{0.7cm}{Mean Filtered Rank}}  \\ 
\hline \hline
13 & \parbox{5.1cm}{$\texttt{plays\_position}^{-1}$  and  $\texttt{plays\_in\_club}^{-1}$ \\
(who plays forward for fc barcelona?)} 
& 94.48 & 1.10 & 71.17 &  1.77 & 87.12 &  1.37 & {\bf 98.77} &  {\bf 1.02} \\ \hline
14 & \parbox{5.1cm}{$\texttt{plays\_position}^{-1}$ and  $\texttt{ plays\_for\_country}^{-1}$ \\
(who are the defenders on german national team?)} 
& 95.93 & 1.08 & 76.42 &  2.50 & 64.23 &  2.02 & {\bf 100.00} & {\bf 1.00} \\  \hline
15 & \parbox{5.1cm}{ $\texttt{ plays\_in\_club}^{-1}$ and $\texttt{ is\_in\_country}^{-1}$ \\
(which player in ssc napoli is from argentina?)} 
& 91.79 & 1.13 & 75.37 &  1.75 & 88.06 &  1.37 & {\bf 94.03} & {\bf 1.07} \\
\hline  \hline
\multicolumn{2}{c|}{Overall} & 94.05 & 1.11 & 74.05 & 1.97 & 80.71 & 1.56 & {\bf 97.62} & {\bf 1.03}
\end{tabular}
}
\end{table}

\section{Knowledge base completion}

\begin{table}[ht]
\caption{Accuracy of knowledge base completion on WordNet. \label{tab:wn-kbc}}
\centering
{\scriptsize
\begin{tabular}[tb]{c|c}
Model & Accuracy (\%) \\
\hline \hline 
TransE (SINGLE) & 68.5  \\
TransE (COMP) & 80.3 \\
\hline
TransGaussian (SINGLE) & 58.4 \\
TransGaussian (COMP) & 76.4
\end{tabular}
}
\end{table}

Knowledge base completion has been a common task for testing knowledge base models on their ability of generalizing to unseen facts.
Here, we apply our TransGaussian model to a knowledge completion task and show that it has competitive performance.

We tested on the subset of WordNet released by \cite{guu2015traversing}.  
The atomic triplets in this dataset was originally created by \cite{socher2013reasoning}
and \cite{guu2015traversing} added path queries that were randomly sampled from the knowledge graph.
We build our TransGaussian model by training on these triplets and paths and tested our model on the same link prediction
task as done by \cite{socher2013reasoning, guu2015traversing}. 

As done by \cite{guu2015traversing}, we trained \emph{TransGaussian (SINGLE)}
with atomic triplets only and trained \emph{TransGaussian (COMP)} with the union of atomic triplets and paths.
We did not incorporate word embedding in this task and each entity is assigned its individual vector.
Without getting parameters tuned too much, \emph{TransGaussian (COMP)} obtained accuracy comparable to 
\emph{TransE (COMP)}. See Table~\ref{tab:wn-kbc}.

\end{document}